\documentclass[letterpaper, 10 pt, conference]{ieeeconf}  

\IEEEoverridecommandlockouts                              

\usepackage{times}
\usepackage{epsfig}
\usepackage{graphicx}
\usepackage{amsmath}
\usepackage{amssymb}
\usepackage{subfig}
\usepackage{tabularx}

\newcolumntype{Y}{>{\centering\arraybackslash}X}

\title{\LARGE \bf
Stretching Domain Adaptation: How far is too far?
}

\author{Yunhan Zhao$^{1}$ \ Haider Ali$^{1}$ \ Rene Vidal$^{1}$ 
\thanks{$^{1}$The Center for Imaging Science,
Johns Hopkins University, Baltimore, MD, 21218, USA}%
}

\begin{document}

\maketitle
\thispagestyle{empty}
\pagestyle{empty}

\begin{abstract}

While deep learning has led to significant advances in visual recognition over the past few years, such advances often require a lot of annotated data. Unsupervised domain adaptation has emerged as an alternative approach that does not require as much annotated data, prior evaluations of domain adaptation approaches have been limited to relatively similar datasets, e.g source and target domains are samples captured by different cameras. A new data suite is proposed that comprehensively evaluates cross-modality domain adaptation problems. This work pushes the limit of unsupervised domain adaptation through an in-depth evaluation of several state of the art methods on benchmark datasets and the new dataset suite. We also propose a new domain adaptation network called ``Deep MagNet'' that effectively transfers knowledge for cross-modality domain adaptation problems. Deep Magnet achieves state of the art performance on two benchmark datasets. More importantly, the proposed method shows consistent improvements in performance on the newly proposed dataset suite.

\end{abstract}


\section{Introduction}

The exponential growth of computational and data infrastructure has enabled deep neural networks to advance the state of the art in visual recognition significantly in just a handful of years. Convolutional Neural Networks (CNNs) have been successfully applied to almost every problem in computer vision, from interest point description~\cite{choy2016nips} to stereo~\cite{zbontar2016jmlr}, object detection~\cite{ren2015nips}, semantic segmentation~\cite{long2015cvpr}, action recognition~\cite{dai2017iccv} and Structure-from-Motion~\cite{clark2017aaai}. However, a key weakness of these approaches is their data-inefficiency. It is not trivial, and often not even possible to collect and label millions of images for a novel task such as robotic navigation, where images may be observed from viewpoints atypical for an internet dataset~\cite{ILSVRC15}, or for a novel sensor where there is little appearance similarity to standard datasets. \\

While it is not uncommon to leverage transfer learning in CNNs when a labeled dataset of appropriate size is available for the target task, only a few datasets and approaches have been proposed for the challenging case of an unlabeled target dataset. Today, unsupervised domain adaptation today is at the stage object recognition was, 15 years ago, with Caltech 101~\cite{fei2004cvpr}. \\

The goal of this work is to stress-test recent domain adaptation approaches by exploiting significantly more challenging datasets. Our objective in doing so is two-fold. First, we wish to uncover the weaknesses of the state of the art methods for this problem. Second, we wish to provide the broader community with baseline results and data to facilitate the development of new ideas. To that end, we conduct an extensive evaluation of existing methods on various domain adaptation datasets of increasing complexity and report results and insights that we expect to be beneficial for future research. Some datasets comprise ``easy'' pairings where both the source and target datasets consist of real centered objects in uncluttered images, Other datasets comprise ``medium'' difficulty pairs where either the source or target dataset consists of crude renderings of 3D CAD models, whereas its counterpart contains real objects in severely cluttered backgrounds. Finally, other datasets comprise ``hard'' pairs where one dataset has human-drawn sketches that typically only contain raw geometry and the objective is to transfer to/from real objects images. We also discuss the performance of Deep MagNet against other state of the art methods and demonstrate its superior performance with respect to cross-modality domain adaptation tasks. \\

Our main contributions can be summarized as follows: 

\begin{itemize}
\item We propose a new dataset suite with the following pairs of source and target sets: (i) randomly 30\% and 70\% samples of real scenes Pascal 3d+~\cite{xiang_wacv14}  (ii) renderings of CAD models, which are instances of $12$ classes obtained from ShapeNet~\cite{shapenet2015}; and real 
scenes from Pascal 3D+ containing the $12$ common categories, (iii) human-drawn coarse sketches, Sketch-250 \cite{sketches} has $87$ common categories in Caltech $256$~\cite{caltech256}.
\item We propose a novel domain adaptation network that outperforms on multimodal domain adaptation problems. Deep MagNet is inspired by~\cite{long2016unsupervised}, where we enforce explicit transfer of knowledge from source to target domains not only at the level of classifier weights but also at the level of convolutional filters. Transition layers are introduced to effectively process the deep network features to make them more adaptable.
\item We conduct a thorough study in terms of classification accuracies comparing multiple state of the art approaches against our proposed network, and report consistently superior or comparable results.
\end{itemize}

\section{Related Work}

As deep neural networks become the state of the art on a variety of visual tasks, there is increasing interest in being able to leverage their capabilities even in the absence of large labeled datasets. Unsupervised domain adaptation approaches aim to transfer the knowledge acquired from labeled datasets, adapting thus learned models on unlabeled datasets. \\

Due to the space constraint, we limit our discussion and categorize work done on the problem into three classes: (i) domain adaptation approaches that directly update the learned weights for a deep neural network on unlabeled data by matching feature distributions~\cite{RevGrad, DAN, long2016unsupervised, haeusser2017iccv, JAN, sohn2017iccv}, (ii) generative approaches~\cite{shrivastava2017cvpr,isola2017cvpr,gan2017} that learn to convert source domain training data to target domain data while retaining labels, and finally (iii) approaches that are not specific to neural networks.\\

\textbf{Matching feature distributions:} This family of approaches~\cite{RevGrad, DAN, long2016unsupervised, haeusser2017iccv, JAN, sohn2017iccv} leverage losses such as the Maximum Mean Discrepancy (MMD) loss~\cite{mmd} to match the statistics of feature activations between layers processing the labeled source datasets and the unlabeled target datasets. The initial studies on semi-supervised domain adaptation~\cite{Gopalan2011} were based on the generative subspaces created from the source and target domains and their underlying properties on benchmark datasets \cite{office31}. Cross-Modal Scene Networks \cite{aytar16} has introduced a variational method to fine-tune a model for task-specific cross model alignment. They have explored which type of models adapts well and which does not as an initial representation in the task of domain adaptation. Additionally, they have introduced the importance of the similar statistical properties of the source and target domains. The primary axes of variation among these approaches include using a different matching loss function used e.g. the so-called ``Walker loss'' proposed by ~\cite{haeusser2017iccv} which outperforms the MMD loss or leveraging adversarial examples as in ~\cite{sohn2017iccv}. We note that our proposed network is orthogonal to these ideas, thus it could still benefit from them which we leave for future work. In particular, We thoroughly evaluate and extend the Residual Transfer Network (RTN) ~\cite{long2016unsupervised}, which applies the MMD loss on classification layer outputs, and also uses an entropy minimization loss on the target domain predictions as an additional regularizer. In addition, RTN learns compositional representations between its source and target function approximations by utilizing extra residual~\cite{he2016deep} blocks for the source classifier. On the other hand, we employ densely connected convolutional networks (DenseNets) ~\cite{huang2016densely} applying the MMD loss not only at the level of classifier activations but also on the convolutional feature learning layers.\\

\textbf{Generative approaches:} Recently, multiple works~\cite{shrivastava2017cvpr,isola2017cvpr,gan2017} leverage conditional Generative Adversarial Networks (cGANs) to convert a labeled source image to a target domain image such that it is indistinguishable from a real target image to a CNN-based discriminator while maintaining the image attributes
relevant to the label. While the images generated in this way can look realistic, the primary focus of these works is not in exploiting this generative knowledge to learn state of the art domain transfer capabilities.\\

\textbf{Non-connectionist approaches:} Significant literature~\cite{TCA, GFK} exists that does not follow the uniform CNN framework for domain adaptation. Some of these papers introduced important domain transfer datasets, including Office-31~\cite{office31} and Office-Caltech~\cite{office-caltech} that we utilize. We compare our approach against many such methods on these standard datasets.

\section{Methodology}

\begin{figure*}[t]
\begin{center}
	\includegraphics[width = \linewidth]{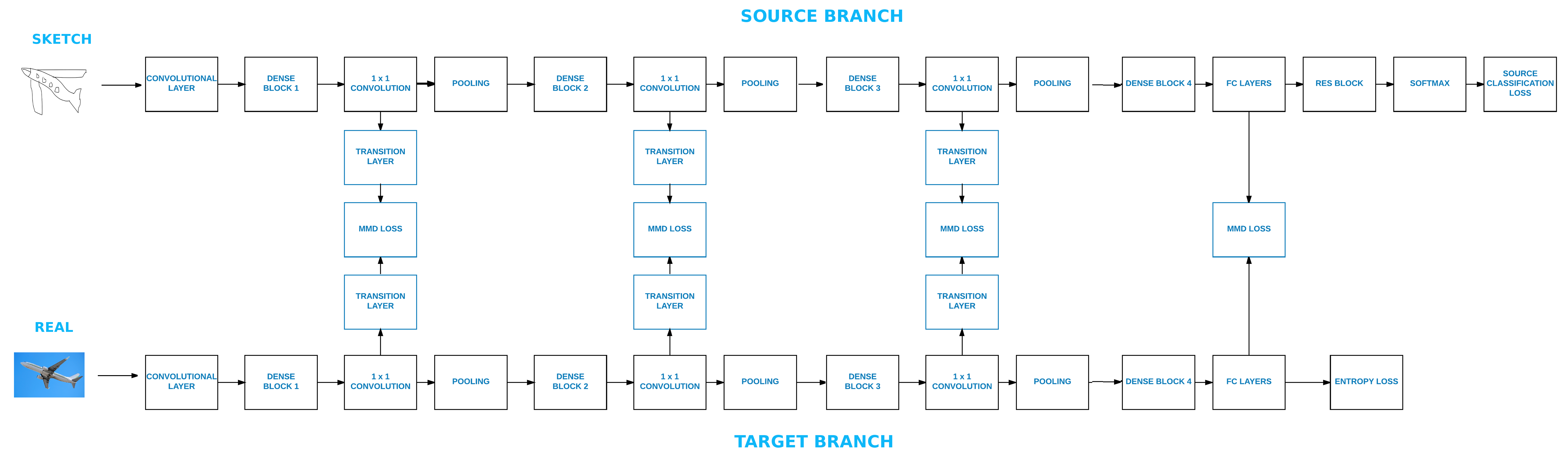}
\end{center}
   \caption{Overview of our proposed ``Deep Magnet'' network architecture.}
\label{fig:network}
\end{figure*}

In the unsupervised domain adaptation problem, we are given a labeled source domain $D_s$ =\{($x_i^s$,$y_i^s)\}_{i=1}^{n_s}$ and an unlabeled target domain $D_t$ = \{$x_j^t\}_{j=1}^{n_t}$, where $x_i^s$ and $x_j^t$ denotes the source and target data respectively, and $y_i^s$ denotes the source labels. The source and target domains are assumed to be sampled from different probability distribution $p$ and $q$, respectively. Our objective is to minimize the expected risk $R_t (f_t) = E_{(x,y) \sim q} [f_t(x) \neq y] $, where $f_t(x)$ is the target domain classifier learned with $D_t$ and $D_s$.	\\

The challenge of unsupervised domain adaptation arises from the fact that the target domain has no labeled data. The source classifier $f_s$, which trained on source domain $D_s$, cannot be directly applied to the target domain $D_t$ due to the distribution discrepancy \cite{discrepancy}. As suggested in \cite{long2016unsupervised}, the source and target classifiers can be connected with a residual block that learns the shortcut connection between two the classifiers. The target classifiers are subsequently represented as a linear combination of the source classifiers. We extend this method to work on cross-modality domain adaptation problems. The source and target features in this type of problems are drastically different from normal domain adaptation problems. Preserving the essential information and effectively aligning source and target features becomes significantly important in cross-modality domain adaptation problems. \\

We propose a network ``Deep MagNet'' that is composed of dense blocks, a residual block (Res Block) and multiple MMD loss functions. Our proposed network is shown in Figure 1. The primary reason for these intermediate layers between dense blocks is to compress the model and shrink the number of parameters in the network. The DenseNet we employed in this network is DenseNet 121 which is one variation of DenseNet. The Res Block is introduced to bridge the source classifiers and target classifiers. There are multiple MMD loss functions in our network to reduce the squared distance between the source and target features at different levels.

\subsection{Transition Layer and Dense Block}

\begin{figure}[ht]
\label{figdense}
\begin{minipage}[c]{0.21\linewidth}
\includegraphics[width=\linewidth]{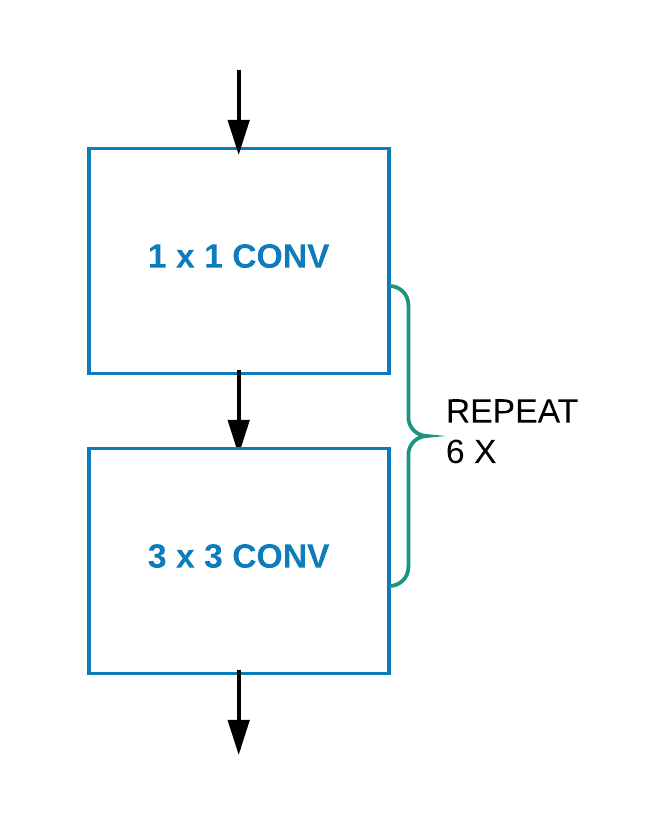}
\tiny
\end{minipage}
\begin{minipage}[c]{0.21\linewidth}
\includegraphics[width=\linewidth]{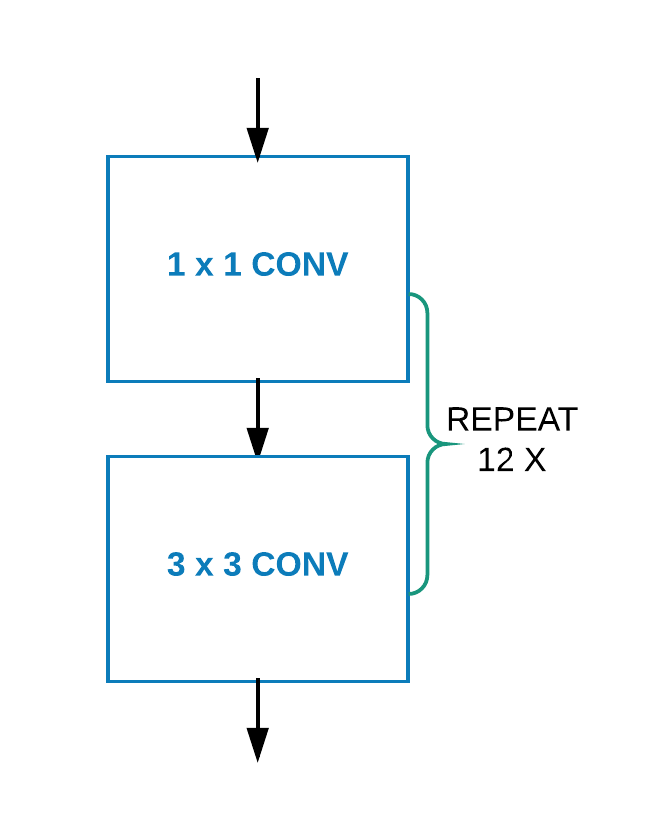}
\tiny
\end{minipage}%
\begin{minipage}[c]{0.2\linewidth}
\includegraphics[width=\linewidth]{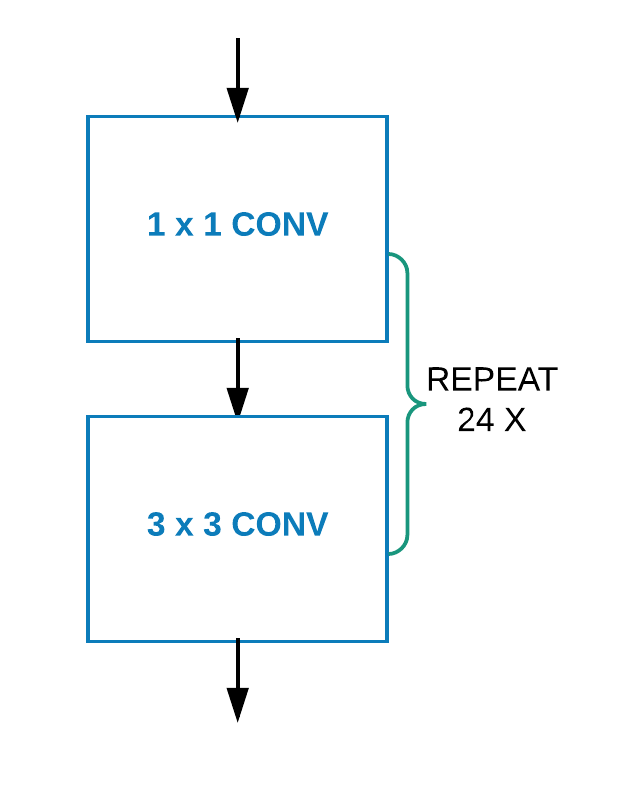}
\tiny
\end{minipage}%
\begin{minipage}[c]{0.22\linewidth}
\includegraphics[width=\linewidth]{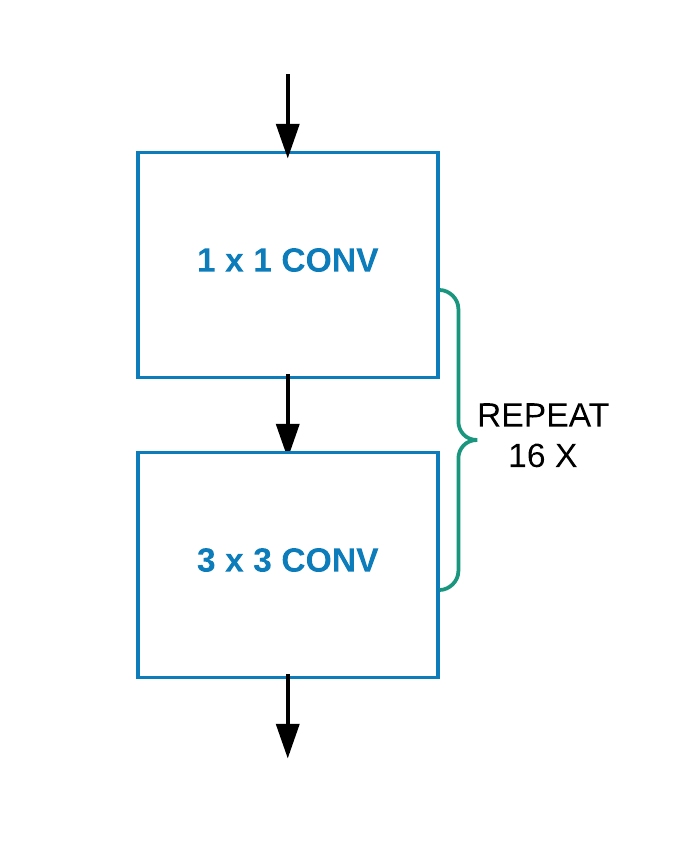}
\tiny
\end{minipage}%
\caption{A simple demonstration of dense blocks. The x here represent the number of repetition of Conv 1$\times$1, 3$\times$3. Each Conv showed in the figure represents the sequence of BN-ReLU-Conv.}
\end{figure}

\begin{figure}[ht] 
    \centering
    \subfloat[Transition Layer A]{%
        \includegraphics[width=0.2\textwidth]{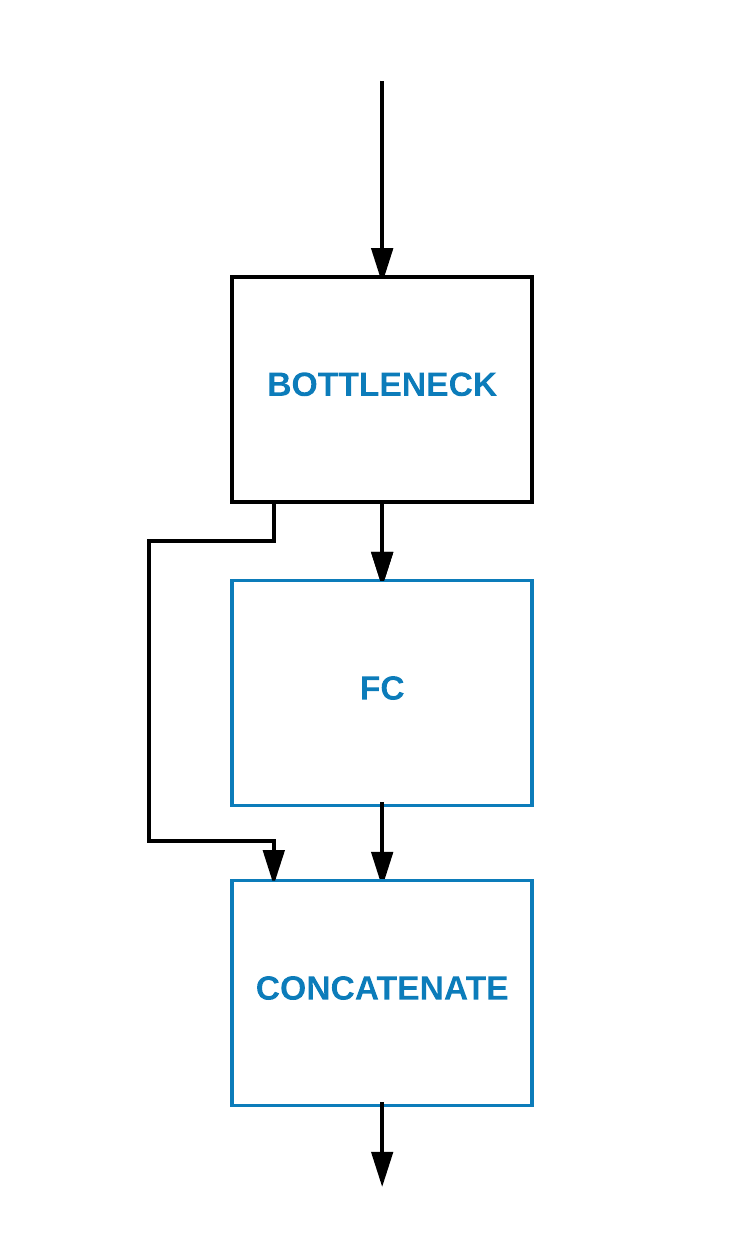}%
        \label{TransitionLayer A}%
        }%
    \hfill%
    \subfloat[Transition Layer B]{%
        \includegraphics[width=0.2\textwidth]{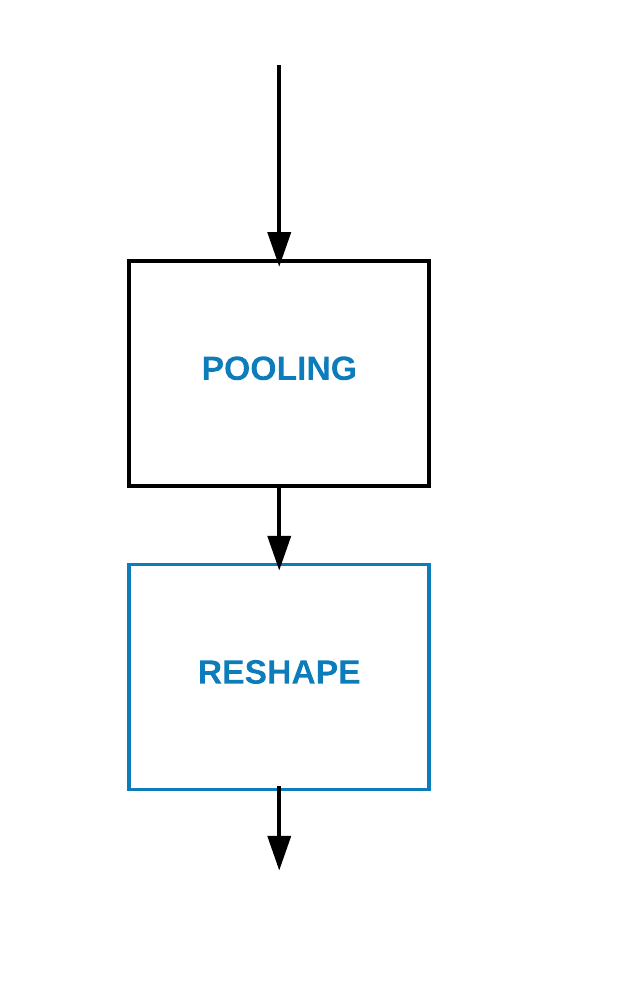}%
        \label{TransitionLayer B}%
        }%
    \caption{Two variations of transition layers applied in our network, where fc means fully connected layers and concatenate represents element-wise addition.}
\end{figure}

Dense blocks described in this paper refer to the unique structure proposed in DenseNet. A sample DenseNet block is illustrated in Figure 2 and all layers are correlated with direct connections within a block. This idea is inspired by attempting maximum information flow between layers in the network. We decide to adopt dense blocks in our network because we believe this structure can capture more structural information than other existing networks. We introduce two variants of transition layers shown in Figure 3 that are designed to accommodate the irregular features extracted by dense blocks. Transition layers are applied after every dense block and process the features into a better state before calculating the distance between features. Transition layers are consistent through the network, which means if we adopt transition layer A in one application then all the transition layers in the network diagram are of type A. Transition layer A involves bottleneck and fully connected layers, which is used on texture informative datasets like Office-31 and Office-Caltech. Transition layer A can further concretize the texture features before comparing them in MMD loss functions. Transition layer B focuses on structural information and it is used in harder datasets like sketches and CAD images. It preserves the shape information and filters out the lower responsive features. 

\subsection{Loss Functions}

The first term of our loss function is the MMD loss, which is widely used in domain adaptation tasks. The features in standard CNNs tend to transition from generic to specific across different layers of the network, and the transferability of features and classifiers across domains will tend to decrease as the discrepancy measure increases. in our design, we use multiple MMD loss functions at different levels of the network to ensure the features between the source and target domains have minimum discrepancy at all level. The MMD loss is defined as:
\begin{equation}
	\begin{aligned}
	MMD(z_s, z_t) &= \sum\limits_i^{n_s} \sum\limits_j^{n_s} \frac{k(z_i^s, z_j^s)}{n_s^2} \\
	&+ \sum\limits_i^{n_t} \sum\limits_j^{n_t} \frac{k(z_i^t, z_j^t)}{n_t^2} \\
	&-2 \sum\limits_i^{n_s} \sum\limits_j^{n_t} \frac{k(z_i^s, z_j^t)}{n_s n_t}
	\end{aligned}
\end{equation}
where $k(\cdot, \cdot)$ represents a kernel function, $z^s$ and $z^t$ represents the source and target features extracted from the network respectively, $n_s$ and $n_t$ represents the total number of samples in the source and target domain, respectively. The kernel function $k(\cdot, \cdot)$ applied in this paper is the standard Gaussian kernel function.	\\

The second term of the loss function is the entropy loss between the source labels and the adaptive classifiers with residual block shown in Figure 1. 
The source classifier and target classifier should be different but they should be correlated to ensure the feasibility of domain adaptation. It is reasonable to assume that $ f_t(x) = f_s(x) + \Delta f(x) $ where the perturbation $ \Delta f(x) $ \cite{perturbation1, perturbation2} is a function of input feature $x$ \cite{he2016deep}. However, this assumption is not optimal because the residual block will make the target classifier similar to source classifier. We want the target classifier to be distinct from the source classifier so that it can best fit the target domain structure. We use the entropy minimization principle to refine the classifier adaptation. More specifically minimizing entropy is solving the following optimization problem:
\begin{equation}
	\min_{w_t} \ \mathbb{E}_{x^t \sim D_t} H(f_t(x^t; w_t))
\end{equation}
where $n_t$ is the number of samples in the target domain, $f_t(\cdot; w_t)$ is the target classifier, $x_i^t$ represents each individual sample from the target domain, $w_t$ is the model parameter. $H$ is the entropy function and it is defined as: 
\begin{equation}
H(f_t(x^t; w_t)) = -\sum\limits_{j=1}^{c} f_t(x^t; w^j_t) \log  f_t(x^t; w^j_t),
\end{equation}
where $c$ is the total number classes and $w_t^j$ is the target classifier weights for the $j^{th}$ class. By minimizing the objective function, the target classifiers are guaranteed to be distinct from the source classifiers. \\

The last part of the loss function is the softmax loss in the source domain and we denote this loss as $L^s$. In summary, our loss function is represented as follows:
\begin{equation}
\begin{aligned}
	L &= \mathbb{E}_{(x^s, y^s) \sim D_s} L^s(f_s(x^s; w_s), y^s)	\\
	&+ \mathbb{E}_{x^t \sim D_t} H(f_t(x^t; w_t))	\\
	&+ \sum_{i=1}^{n_{MMD}} \mathbb{E}_{(z_s, z_t) \sim (f_s(D_s; w_s), f_t(D_t; w_t))} \ MMD(z_s, z_t)
\end{aligned}
\end{equation}
where $w_s$ is the model parameter of function $f_s$ and $n_{mmd}$ is the total number of MMD loss functions adopted in the network. $L^s$ is the negative log-likelihood function and it is written as:
\begin{equation}
    L^s(f_s(x^s; w_s), y_s) = - \log p(y^s | x^s; w_s),
\end{equation}
Under multi-class condition, the probability $p(y^s | x^s)$ is represented as:
\begin{equation}
    p(y^s = k | x^s) = \frac{e^{f_s(x^s; w^k_s)}}{\sum\limits_{i=1}^c e^{f_s(x^s; w^i_s)}}
\end{equation}
The softmax loss function and the entropy loss function are applied at the end of the network. As shown in the network architecture, MMD loss functions are applied at each intermediate layers between dense blocks and one more MMD loss function is applied at the end of the network.

\section{Experiments}

In this section, we will introduce the experimental setup that includes our implementation details,  benchmark datasets and the new dataset suite we proposed. We validate our method on benchmark datasets like Office-31 and Office-Caltech, then we implement our network and other baselines on the new dataset suite. We follow standard evaluation protocols for unsupervised domain adaptation problems. We evaluate our method and other baselines based on average classification accuracy and compare the average classification accuracy of different methods on the same transfer tasks. The discussion of results is at the end of this section. \\

\begin{figure}[ht]
\includegraphics[width=\linewidth]{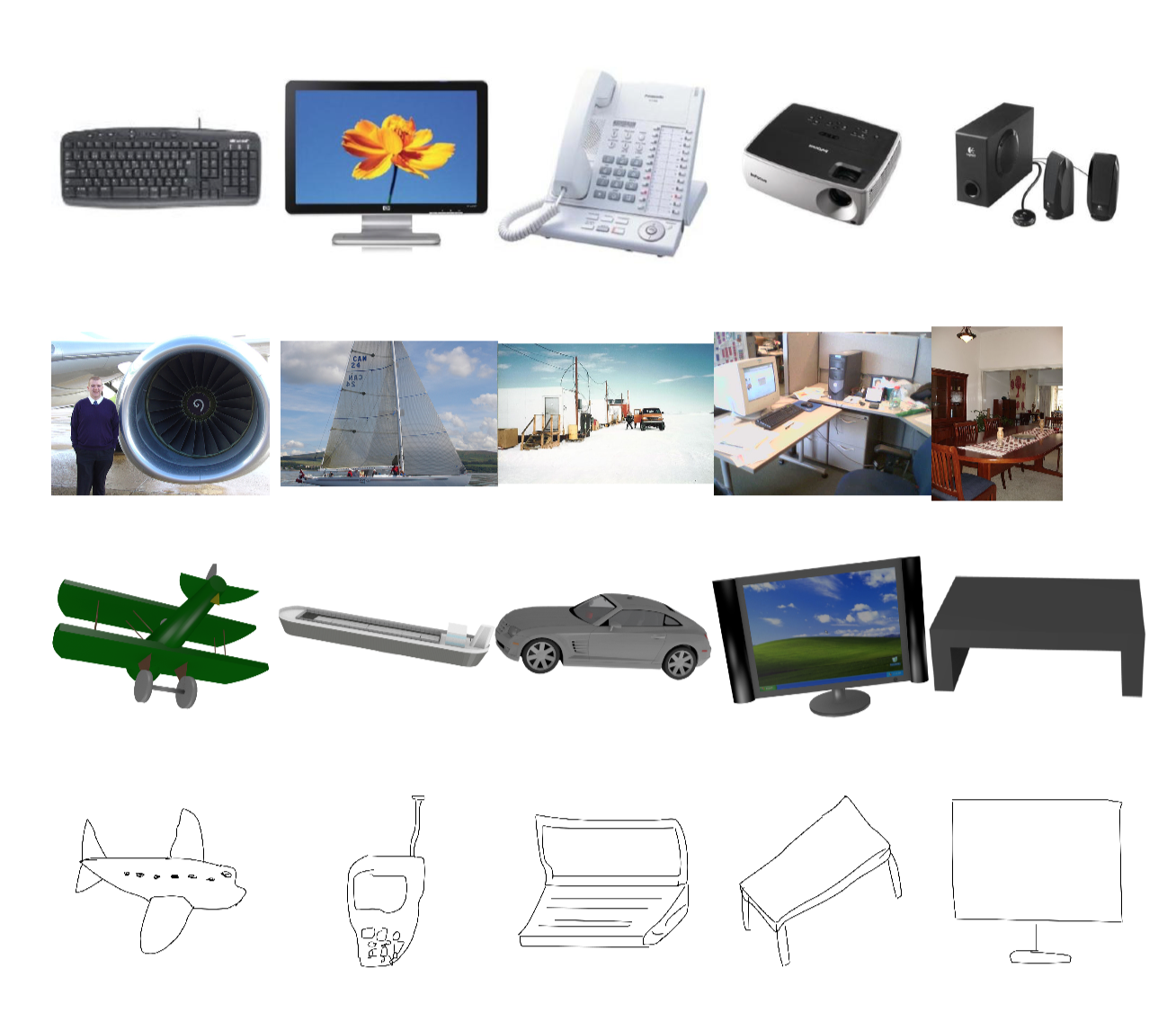}
\caption{Each row from top to bottom represents image samples from Office-31, Pascal 3D+, ShapeNet and Sketch-250, respectively.}
\label{fig: sampleData}
\end{figure}

We conducted all our experiments on the Caffe deep learning framework \cite{caffe} and fine-tuned our models from the author provided pre-trained Caffe models. More specifically, we fine-tuned Deep MagNet from the DenseNet 121 pre-trained model on all experiments. Other methods follow their original implementations unless specified otherwise. All the numbers reported in this section are the average of multiple repetitions. \\

We have conducted a grid search on learning rate and it turns out that 0.003 shows the best performance. Thus, we set the basic learning rate to be 0.003 for layers that learn from scratch and 10 times higher for layers that have pre-trained weights. We fed two type of features into MMD loss layers in order to accommodate different domain adaptation problems. One is fully connected features that correspond to transition layer type A and the other is convolutional features that correspond to transition layer type B. Fully connected features are much more specific than convolutional feature so we utilize fully connected features when we want the neural networks to learn low-level information and we use convolutional features when we want neural network to focus more on high-level features. We apply the initialization methods introduced in \cite{init}. The optimizer is mini-batch Nesterov's Accelerated Gradient Descent \cite{Nesterov} with a momentum of 0.9 and the learning rate annealing strategy \cite{RevGrad} is implemented. Learning rate annealing strategy can self-adjust the learning during the optimization, which stabilizes the outputs and potentially boosts the performance of networks. We run all experiments with 50,000 iterations after numerous empirical tests. Experimental results show that the accuracy saturates before 50,000 iterations. Thus, we decide to stop the experiments at 50,000 iterations to save computational resources and prevent potential overfitting.

\subsection{Real to Real Images}

\subsubsection{Office-31 and Office-Caltech Experiments}
We first evaluate our network with two benchmark datasets: Office-31 and Office-Caltech. Office-31 is a standard benchmark dataset of domain adaptation problems, including 4,652 images of 31 categories collected from three different domains: Amazon (A), which contains images downloaded from amazon.com, Webcam (W), which includes images taken by web camera, DSLR (D), which holds images taken by a digital SLR camera. Office-Caltech is an extension of the Office-31 dataset and it is built by selecting the 10 common categories shared between Office-31 and Caltech-256 (C). Office-Caltech has one more domain than Office-31 but much fewer images in each domain. Office-Caltech is still a great alternative to check the robustness of the domain adaptation algorithms. We build transfer tasks between all domains for both datasets.

\begin{table*}[ht]
\begin{center}
\begin{tabularx}{\textwidth}{| l |Y|Y|Y|Y|Y|Y|Y|}
\hline
Method & A $\rightarrow$ W & D $\rightarrow$ W & W $\rightarrow$ D & A $\rightarrow$ D & D $\rightarrow$ A & W $\rightarrow$ A  & Avg \\
\hline\hline
TCA & 59.0\% & 90.2\% & 88.2\% & 57.8\% & 51.6\% & 47.9\% & 65.8\% \\
GFK & 58.4\% & 93.6\% & 91.0\% & 58.6\% & 52.4\% & 46.1\% & 66.7\% \\
AlexNet & 60.6\% & 95.4\% & 99.0\% & 64.2\% & 45.5\% & 48.3\% & 68.8\% \\
DDC & 61.0\% & 95.0\% & 98.5\% & 64.9\% & 47.2\% & 49.4\% & 69.3\%	\\
DAN & 68.5\% & 96.0\% & 99.0\% & 66.8\% & 50.0\% & 49.8\% & 71.7\%	\\
CORAL & 66.4\% & 95.7\% & 99.2\% & 66.8\% & 52.8\% & 51.5\% & 72.1\% \\
RTN	& 73.3\% & 96.8\% & 99.6\% & 71.0\% & 50.5\% & 51.0\% & 73.7\%	\\
Compact DNN & 82.6\% & 97.0\% & 99.4\% & 80.1\% & 67.3\% & 67.4\% & 82.3\% \\
JAN & \textbf{85.4\%} & 97.4\% & 99.8\% & \textbf{84.7\%} & 68.6\% & \textbf{70.0\%} & \textbf{84.3\%} \\
\hline
Deep MagNet & 83.5\% & \textbf{98.3\%} & \textbf{100.0\%} & 82.5\% & \textbf{70.4\%} & 69.7\% & 84.1\% \\
\hline
\end{tabularx}
\end{center}
\caption{The average classification accuracy on the Office-31 dataset, where A, W, D represent Amazon, Webcam and DSLR, respectively.}
\label{tab: 1}
\end{table*}

\begin{table*}[ht]
\begin{center}
\begin{tabularx}{\textwidth}{|l| Y| Y| Y| Y| Y| Y| Y| Y| Y| Y| Y| Y| Y |}
\hline
Method & A$\rightarrow$W & D$\rightarrow$W & W$\rightarrow$D & A$\rightarrow$D & D$\rightarrow$A & W$\rightarrow$A  & A$\rightarrow$C & W$\rightarrow$C & D$\rightarrow$C & C$\rightarrow$A & C$\rightarrow$W & C$\rightarrow$D & Avg \\
\hline\hline
TCA & 84.4\% & 96.9\% & 99.4\% & 82.8\% & 90.4\% & 85.6\% & 81.2\% & 75.5\% & 79.6\% & 92.1\% & 88.1\% & 87.9\% & 87.0  \\
GFK & 89.5\% & 97.0\% & 98.1\% & 86.0\% & 89.8\% & 88.5\% & 76.2\% & 77.1\% & 77.9\% & 90.7\% & 78.0\% & 77.1\% & 85.5 \\
AlexNet & 79.5\% & 97.7\% & 100.0\% & 87.4\% & 87.1\% & 83.8\% & 83.0\% & 73.0\% & 79.0\% & 91.9\% & 83.7\% & 87.1\% & 86.1 \\
DDC & 83.1\% & 98.1\% & 100.0\% & 88.4\% & 89.0\% & 84.9\% & 83.5\% & 73.4\% & 79.2\% & 91.9\% & 85.4\% & 88.8\% & 87.1	\\
DAN & 91.8\% & 98.5\% & 100.0\% & 91.7\% & 90.0\% & 92.1\% & 84.1\% & 81.2\% & 80.3\% & 92.0\% & 90.6\% & 89.3\% & 90.1	\\
RTN	 & 95.2\% & 99.2\% & 100.0\% & 95.5\% & 93.8\% & 92.5\% & 88.1\% & 86.6\% & 84.6\% & 93.7\% & 96.9\% & 94.2\% & 93.4	\\
Compact DNN & 95.6\% & 99.7\% & 100.0\% & 96.8\% & \textbf{96.0}\% & \textbf{95.6}\% & 92.5\% & 91.6\% & \textbf{91.4}\% & 95.3\% & 97.2\% & \textbf{95.3}\% & 95.6 \\
\hline
Deep MagNet & \textbf{97.2}\% & \textbf{99.7}\% & \textbf{100.0}\% & \textbf{97.2}\% & 95.8\% & 95.2\% & \textbf{93.8}\% & \textbf{92.2}\% & 90.3\% & \textbf{96.2}\% & \textbf{98.6}\% & 95.0\% & \textbf{95.9} \\
\hline
\end{tabularx}
\end{center}
\caption{The average classification accuracy on the Office-Caltech dataset, where C represent Caltech-256 and A, W, D represent the same domains as Table \ref{tab: 1}.}
\label{tab: 2}
\end{table*}

\subsubsection{Pascal 3D+ Experiments}

Even though Office-31 and Office-Caltech are widely used, they all have certain limitations that cannot fully evaluate real to real images problems. The number of images of different domains in Office-31 might have significant differences, which makes the difficulty of some transfer tasks much higher others. Therefore, we have observed that some transfer tasks are already saturated while others are relatively hard to improve due to the insufficient data samples. Meanwhile, the Office-Caltech dataset is a relatively small dataset, which makes the results less convincing when neural networks are getting deeper and deeper. \\

Due to the reasons above, we introduce a new experiment in this section to complement the weakness of Office-31 and Office-Caltech. Pascal 3D+ is a publicly available dataset released by Stanford University. This dataset contains CAD models and real images of 12 common categories. In this section, we only use the real images of this dataset. We split this dataset into the training set and test set by randomly selecting 70\% and 30\% of the images from each category, which means the training and test sets still have the same distribution as the original dataset. The idea behind this experiment is to test the performance of networks on real images from a different perspective since Office-31 and Office-Caltech datasets are not considered as natural images. Images in both Office datasets have nicely centered objects and images either do not have any background or have very similar backgrounds. Real images in Pascal 3D+ dataset are natural images, which is a great complement to stretch the idea of real to real modality. Pascal 3D+ images have possibly more objects in one single image and more variations in terms of the background. Moreover, the size of the object in the image and the viewing angle of the object vary across images.

\subsection{CAD to Real Experiments}
\label{cadtorealexperiments}

We evaluate our approach on simple cross-modality domain adaptation problems, which means transferring between CAD and real images. CAD images that mentioned in this section are rendered images of CAD models from ShapeNet. These images are generated by rotating and resizing each CAD model with different magnitudes and project them onto a blank background. CAD rendered images contain full or most likely partial shape information but they might have drastic differences in color or texture compared with normal RGB images. The number of rendered CAD images generated are around two million and we build the final ShapeNet dataset by uniformly sampling 20\% of the rendered images from each category. We find that 20\% is already large enough for CNNs to learn useful features after several experiments and this also reduces the computational cost. The corresponding real dataset is built by choosing 12 common categories from the Pascal 3D+ dataset. The total number of images in our ShapeNet dataset is 108,000 and the total number of images in the real dataset is 30,362. We also conduct a reverse experiment that performs domain adaptation task: Pascal 3D+ $\rightarrow$ ShapeNet. The purpose of this experiments is to further validate the ability of networks to learn from real images and apply knowledge on CAD.

\begin{table*}[ht]
\begin{center}
\begin{tabularx}{\textwidth}{|l | Y | Y | Y | Y | Y | Y|}
\hline
Methods & P $\rightarrow$ SN & SN $\rightarrow$ P & S $\rightarrow$ C & C $\rightarrow$ S & P70 $\rightarrow$ P30 & Avg \\

\hline\hline
AlexNet & 39.4\% & 45.2\% & 25.7\% & 12.6\% & 83.0\% & 41.18\% \\
DenseNet-121 & 52.1\% & 57.4\% & 34.1\% & 10.2\% & 88.5\% & 48.46\% \\
ResNet-152 & 53.8\% & 64.4\% & 52.8\% & 14.1\% & 88.7\% & 54.76\% \\
CORAL & 41.4\% & 55.4\% & 35.8\% & 37.8\% & 84.7\% & 51.02\% \\
RTN & 42.4\% & 65.7\% & 16.8\% & 18.6\% & 84.2\% & 45.54\% \\
JAN(AlexNet) & 51.2\% & 63.5\% & 42.1\% & 33.1\% & 84.9\% & 54.96\%  \\
JAN(ResNet-152) & 57.6\% & 67.5\% & 46.8\% & 48.5\% & 87.9\% & 61.66\%  \\
\hline

Deep MagNet & \textbf{58.3\%} & \textbf{68.5\%} & \textbf{53.5\%} & \textbf{50.5\%} & \textbf{89.2\%} & \textbf{64.00\%} \\
\hline
\end{tabularx}
\end{center}
\caption{Average classification accuracy of cross-modality domain adaptation problems. S represents Sketch-250, SN represents ShapeNet, P represents Pascal 3D+, C represents Caltech-256, P70 and P30 represents 70\% and 30\% of Pascal 3D+, respectively.}
\label{tab: 3}
\end{table*}

\subsection{Sketches to Real Experiments}

Now we want to push domain adaptation problems to an even harder level by transferring between human-drawn sketches and real images. The Sketch-250 dataset is a dataset with 250 categories of human-drawn sketches as shown in the last row of Figure 4. The objects in human-drawn sketches are very likely to have missing parts due to the knowledge transfer limitation of the person who makes the sketches. Furthermore, the raw shape information contained in sketches is much less accurate than real and CAD rendered images. Caltech-256 is a publicly available dataset that has 256 categories of daily objects and it shares 87 common categories with Sketch-50. We build the transfer task with Sketch-250 as source domain and Caltech-256 as target domain. The source domain, Sketch-250 has 6960 images of 87 categories and the target domain, Caltech-256 has 11712 images of 87 categories. We conduct one more experiment by exchanging source and target domains just like in the CAD to real images section. We have accomplished the experiment: Caltech-256 $\rightarrow$ Sketch-250, which is consistent with experiments in CAD to real images section.

\subsection{Baselines}

We compare against some conventional methods, like TCA \cite{TCA}, GFK \cite{GFK}, DDC \cite{DDC} and DAN \cite{DAN}, as well as state of the art methods like RTN, CORAL, Compact DNN \cite{compact} and JAN \cite{JAN} on Office-31 and Office-Caltech datasets. We first compare against classic neural networks trained on ImageNet, like AlexNet, ResNet and DenseNet on new dataset suite. Moreover, we also include other state of the art domain adaptation methods into our comparison in cross-modality transfer tasks, e.g. RTN, CORAL, Compact DNN and JAN. These methods represent different perspectives of current state of the art approaches in domain adaptation. Comparing against a diverse state of the art methods greatly help us evaluate the performance of the proposed Deep MagNet.

\subsection{Results}

The experimental results of the six real to real transfer tasks of Office-31 are shown in Table \ref{tab: 1}. Our proposed method performs better in some tasks and comparable to state of the art methods on average. The result of twelve real to real transfer tasks of Office-Caltech are shown in Table \ref{tab: 2}. Deep MagNet demonstrates very promising results on this dataset and outperforms other state of the art methods. Our network has the best performance on real to real transfer task on the newly proposed data suite as shown in Table \ref{tab: 3}. Deep MagNet has the state of the art performance on not only classic datasets but also on the proposed dataset suite. Note that real images in Pascal 3D+ do not have well-defined spatial structures. The location and scale of objects in the images are highly unpredictable, which makes this transfer task harder than Office-31 and Office-Caltech datasets. Methods that heavily rely on texture information are very likely to achieve great performance on Office-31 and Office-Caltech but not on Pascal 3D+. The images in Pascal 3D+ require more varied features to achieve better classification results, which fully demonstrates the diverse features learned by Deep MagNet.  \\

The average classification accuracy of transfer tasks between CAD and real images are shown in Table \ref{tab: 3}, e.g. $P \rightarrow SN$ and $SN \rightarrow P$. Domain adaptation tasks between CAD and real are harder than real to real images problems. CAD rendered images lack texture information, which creates an information mismatch between source and target domains. However, CAD rendered images and RGB images still share many features in common, such as the shape of objects. Therefore, transferring between CAD and real images requires methods to capture the high-level features between the source and target domains. Deep MagNet consistently outperforms other state of the art methods on this transfer task. We can also observe that state of the art real methods that outperforms on real to real transfer tasks have relatively less desired performance on CAD to real problems. The performance of Deep MagNet on this task is better than on real to real transfer tasks on Office-31 and Office-Caltech datasets. \\

The average classification accuracy of transfer tasks between sketches and real images are shown in Table \ref{tab: 3}, e.g. $S \rightarrow C$ and $C \rightarrow S$. The domain adaptation problems between sketches and real images are even harder than CAD to real images. CAD rendered images preserve accurate shape information while sketches only provide rough shapes and partially missing objects. The inaccurate shape and missing information make neural networks struggle to learn effective features, which make the transfer tasks harder than previous two type of transfer problems. The experimental results also prove this idea since the average classification accuracy of Sketch-250 and Caltech-256 is the lowest among all experiments. Deep MagNet has the best performance on this extremely hard domain adaptation tasks.

\subsection{Discussion}

We can observe that the average classification results of AlexNet, DenseNet-121 and ResNet-152 on the new dataset suite meet our expectation. The deeper models generally have higher classification accuracy under the same conditions. When we compare other state of the art domain adaptation methods, like RTN,  CORAL and JAN, it is easy to observe that they have better results than networks that are trained on the ImageNet, e.g. AlexNet, ResNet and DenseNet. State of the art domain adaptation methods show great ability in transferring between real images with structured spatial location. But their performance degrades as the transfer tasks become harder. Deep MagNet has a more versatile and flexible transfer ability since it preserves constant outstanding performance on all transfer tasks presented in the experiment section. \\

Deep MagNet has significantly better transfer ability and generalization ability than RTN. RTN shows great performance on Office-31 and Office-Caltech datasets while it struggles on the newly proposed dataset suites. RTN employs MMD loss functions that compare the source features and target features at the end of the network, which provides less flexibility in some cases. Office-31 and Office-Caltech datasets have well-defined spatial information, therefore feed high-level features into MMD loss functions boost the performance. This approach is expected to have performance degradation when it meets more challenging tasks where the images in source and target domains are significantly different. Our method extends RTN and overcomes the over-reliance on the pixel level information, which proves the advantages of having multiple MMD loss functions and transition layers. Shape information is not formed in the last layer since it is developed along the network. Multiple MMD loss functions will compare source and target features at different levels of the network to ensure all essential information preserved. Additionally, we attempted to replace the pre-trained models of RTN, e.g. replace AlexNet or ResNet with DenseNet while keeping everything else same. The results show that they have less desired performance compared to their original structures. The transition layers solved this problem successfully since Deep MagNet adopts dense blocks but still maintain promising performances. Dense blocks produce drastically different features and the transition layers make these features become adaptable to MMD loss functions. \\

Given the results in Table \ref{tab: 3}, we believe the newly proposed data suite validate the transfer ability from a broader perspective. This new data suite validates not only real to real images transfer tasks but also real to CAD rendered images and real to sketches transfer tasks. The three distinct types of transfer tasks contained in the dataset suite provide the sufficient validation of the quality of the domain invariant features. 

\section{Conclusion}
This paper proposed a new dataset suite that effectively evaluates the ability to transfer knowledge between significantly different domains. A thorough evaluation of current state of the art methods on new dataset suited is presented in this paper. The state of the art methods show struggling performance on the cross-modality domain adaptation challenges. Deep MagNet is proposed and it shows consistent good performance on both conventional datasets and the new dataset suite. A transition layer is introduced to process the features, which minimizes the loss function more effectively.

{\small
\bibliographystyle{ieeetr}
\bibliography{root}
}

\end{document}